# Noisy-OR Models with Latent Confounding

**Antti Hyttinen**
HIIT & Dept. of Computer Science
University of Helsinki
Finland

**Frederick Eberhardt**
Dept. of Philosophy
Washington University in St. Louis
Missouri, USA

**Patrik O. Hoyer**
HIIT & Dept. of Computer Science
University of Helsinki
Finland

## Abstract

Given a set of experiments in which varying subsets of observed variables are subject to intervention, we consider the problem of identifiability of causal models exhibiting latent confounding. While identifiability is trivial when each experiment intervenes on a large number of variables, the situation is more complicated when only one or a few variables are subject to intervention per experiment. For *linear* causal models with latent variables Hyttinen et al. (2010) gave precise conditions for when such data are sufficient to identify the full model. While their result cannot be extended to discrete-valued variables with *arbitrary* cause-effect relationships, we show that a similar result can be obtained for the class of causal models whose conditional probability distributions are restricted to a 'noisy-OR' parameterization. We further show that identification is preserved under an extension of the model that allows for negative influences, and present learning algorithms that we test for accuracy, scalability and robustness.

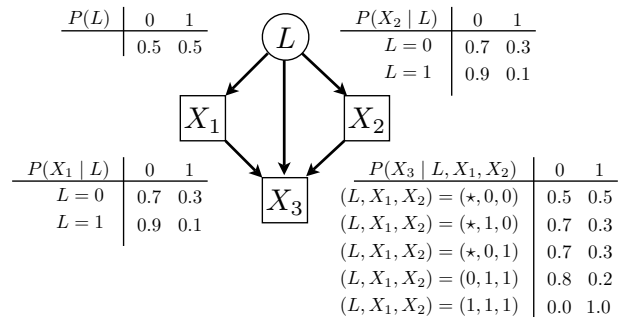

Figure 1: A simple causal model with observed variables $X_1$, $X_2$, $X_3$ and a latent confounder $L$. The '$\star$'-symbol in the conditional probability tables is a wild card that denotes either value. See main text and Appendix A for details and discussion.

## 1 INTRODUCTION

In many data-analysis situations one would like to understand a system of complex interactions between some set of variables of interest. Modeling the statistical relationships among the observed variables in a non-experimental (i.e. 'passive observational') setting is sufficient to predict the behavior of the system in its natural state, but if our ultimate aim is to control the system, or to predict how the variables would respond to outside influence, we need to model the *causal* relationships between the variables.

Here, we restrict our attention to causal systems represented by directed acyclic graphical models (Spirtes et al., 1993; Pearl, 2000). In such a model, a directed acyclic graph (DAG) is used to represent the direct causes among the variables, while an associated set of parameters defines the specifics of the dependence of each variable on its direct causes. An example is given in Figure 1, where the square nodes represent the three observed variables, the circular node denotes a hidden (latent) variable, and directed arcs denote the existence and direction of direct cause-effect relationships. The associated conditional probability tables indicate the probability distribution of each node conditioned on the values of its parents.

We are interested in approaches by which such causal models can be inferred from data. Given a non-experimental dataset, one prominent approach to structure learning is based on identifying statistical independencies in the data (Pearl, 2000; Spirtes et al., 1993). While some methods make an explicit assumption that there are no hidden variables, there exist procedures that take into account the possibility of latent confounding (Spirtes et al., 1995). Unfortunately, the output of such procedures is often quite uninformative when there is a fair amount of confounding. For instance, in a passive observational distribution the model in Figure 1 entails no conditional or marginal independencies among the observed variables, hence independence-based approaches would not yield any substantial causal conclusions in this case.

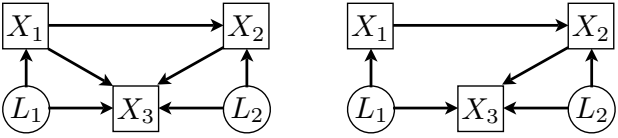

Figure 2: Two graphs differing only in the direct effect from $X_1$ to $X_3$. Given only passive-observational data and data from experiments randomizing a single variable, the existence of the direct effect is underdetermined.

Given that passive observational data often leave the causal relations severely underdetermined, one typically would like to use data from randomized controlled experiments. In such experiments, one or several of the observed variables are randomized, breaking the influence of the natural causes of the variable(s) in question. This has the obvious benefit that any correlation between an intervened variable and another variable can unambiguously be attributed to a causal effect of the former on the latter. It is easy to see that if we are able to intervene on *all but one* of the observed variables, we can use standard statistical methods to estimate the direct causal relationships from all the intervened variables to the remaining observed variable. Consequently, for $n$ observed variables a set of $n$ such all-but-one experiments can be used to determine the entire structure of the underlying graph.

Unfortunately, experiments intervening on a large number of variables at a time are often expensive or otherwise infeasible, so practitioners may have to rely on experiments that intervene on only one or a few of the variables at a time. In general, such experiments are *insufficient* to identify the complete graphical model. In Appendix A we show how the example of Figure 1 is such a case and could lead to incorrect inferences: Given only passive observational data and the responses to experiments randomizing a single variable, one might infer that in order to obtain a high probability of $X_3 = 1$, the optimum two-variable intervention on $X_1$ and $X_2$ would be to set both to zero. However the true optimum is to set both to one.

The above problem is not restricted to inferences about the parameterization, but generalizes to an underdetermination of the causal structure: While experiments randomizing a single variable will typically reveal whether a given variable has a causal effect on another, such experiments do not indicate whether that effect is *direct* or merely *indirect* (mediated by one or more of the other observed variables). For instance, the two graphs of Figure 2 yield the same pattern of statistical dependencies in both the passive observational case and for all experiments randomizing a single variable. Hence, without further assumptions, procedures based on independence tests alone, such as the FCI- and MCI-algorithms (Spirtes et al., 1995; Claassen and Heskes, 2010), cannot in general be used to distinguish causal structures such as these even if there are experiments. When the underlying causal relations can take any form this underdetermination is not limited to procedures based on independence tests: There exist parametrizations for the two graphs of Figure 2 that yield the exact same distributions in both the passive observational case and for all experiments randomizing a single variable, so that the existence of the direct effect cannot be determined unless *both* $X_1$ and $X_2$ are subject to an intervention in the same experiment.

However, if the parametric form specifying the dependence of each variable on its direct causes is constrained, full model identification is possible in some cases. Recently, Eberhardt et al. (2010) and Hyttinen et al. (2010) considered the case of continuous-valued variables $X_j$ ($j = 1, \ldots, n$) related by *linear* structural equations

$$X_j := \sum_{i \in \mathrm{Pa}(j)} b_{ij} \cdot X_i + E_j, \qquad (1)$$

where $\mathrm{Pa}(j)$ denotes the set of parents of variable $X_j$, the parameters $b_{ij}$ represent the strength of the influence of $X_i$ on $X_j$, and the $E_j$ ($j = 1, \ldots, n$) are stochastic disturbance variables. Unobserved confounding (due to latent variables) was in this model implicitly represented by allowing arbitrary covariances among the disturbance variables. Hyttinen et al. (2010) proved that if a set of experiments contains for each ordered pair of observed variables $(X_i, X_j)$ one experiment such that $X_i$ is subject to intervention and $X_j$ is observed, and one experiment in which both are observed, then the model is fully identifiable. This so-called *pair condition* was shown to be both necessary and sufficient for identifiability.

Here we consider parametrizations of causal graphical models for *binary* random variables *with latent confounding*, for which identifiability can be guaranteed. In Section 2 we give a formal definition of such a model based on a 'noisy-OR' parameterization of the conditional probability distributions. While we found that the method of identification had to be very different from the continuous-linear case discussed above, we show in Section 3 that our model is identifiable for con-ditions that are essentially the same. In Section 4 we describe two learning algorithms; the first is based on the efficient selection of conditioning sets, the second uses the EM-algorithm to provide maximum likelihood estimates. While the basic model can only represent systems in which variables have positive influences on each other, we broaden the model class in Section 5 to allow for negative effects, while retaining identifiability. Finally, the effectiveness of the model and algorithms are tested in Section 6.

## 2 NOISY-OR MODEL WITH LATENT CONFOUNDING

We begin by considering causal models for binary variables, in which the dependence of each variable as a function of its parents takes a simple noisy-OR parametrization (Pearl, 1986; Peng and Reggia, 1986). The value of a variable $X_j$ is defined as a function of its parents (direct causes) $X_{\text{Pa}(j)}$, a set of binary 'link' random variables $B_{ij}$, and a binary 'disturbance' random variable $E_j$, by the structural equation:

$$X_j := \bigvee_{i \in \text{Pa}(j)} (B_{ij} \wedge X_i) \vee E_j, \qquad (2)$$

where $\wedge$ and $\vee$ denote binary AND and OR, respectively. Note the similarities to the linear model of (1). In the standard noisy-OR model, all the $E_j$'s and the $B_{ij}$'s are mutually independent. Such a model can also be represented as a Bayesian network without hidden variables (i.e. no confounding). The probability distribution over the observed variables $X_1^n = (X_1, \ldots, X_n)$ is given by

$$P(X_1^n \mid E_1^n) = \prod_{j=1}^{n} P(X_j \mid X_{\text{Pa}(j)}, E_j), \qquad (3)$$

with  $P(X_j = 1 \mid X_{\text{Pa}(j)}, E_j) = \qquad (4)$
$$1 - (1 - E_j) \prod_{i \in \text{Pa}(j)} (1 - b_{ij})^{X_i},$$

where  $b_{ij} = P(B_{ij} = 1)$  and  $P(E_1^n) = \prod_{j=1}^{n} P(E_j)$.

As in linear structural equation models, the direct causal relationship between any pair of variables is defined by just one parameter $b_{ij}$. However, the similarity has its limits: In the linear case the identifiability results derive primarily from the fortunate circumstance that correlations between variables can be computed by so-called 'trek rules'. That is, the correlation between $X_i$ and $X_j$ is given by the sum-product of all the edge-coefficients $b_{(..)}$ along all the causal treks connecting $X_i$ and $X_j$. Danks and Glymour (2001) showed that these trek rules for correlations do not generalize to structures with a noisy-OR parameterization, unless the variables are singly connected.

Instead, a different property of the noisy-OR parameterization turns out to be useful for discovery: *context specific independence*. To specify this property in the most general terms, we use double bars to denote variables that are subject to a randomization and single bars to denote standard conditioning. If both occur in one term, then the conditioning occurs with respect to the interventional distribution. With this notation, we can now define the context specific independence of two variables $X_{i_1}$ and $X_{i_2}$ with common child $X_j$ as

$$(X_{i_1} \perp\!\!\!\perp X_{i_2} \mid X_C \parallel X_I) \qquad (5)$$
$$\Rightarrow (X_{i_1} \perp\!\!\!\perp X_{i_2} \mid X_j = 0, X_C \parallel X_I),$$

where $X_I$ and $X_C$ are (possibly empty) sets of variables and $X_j \notin X_C$. In words: If two variables are conditionally independent in some context specified by $X_C$ and $X_I$, then additionally conditioning on their common child $X_j = 0$ does not destroy the independence. The context is given by conditioning on the variables $X_C$ in the interventional distribution resulting from independent and simultaneous interventions on the variables in the set $X_I$. This independence property is evident from Equation 2: $X_j = 0$ already implies that $(B_{i_1 j} \wedge X_{i_1}) = 0$, so the specific value of the other cause $X_{i_2}$ does not provide any additional information about $X_{i_1}$. Note that the above does not hold when conditioning on $X_j = 1$.

We still have to consider how to represent latent variables (confounding). Generally the causal structure among the latents may be complicated, but the resulting effect on the observed variables can still be quite simple. Thus, instead of representing latents explicitly, we model confounding by allowing the 'disturbances' $E_1^n = (E_1, \ldots, E_n)$ to be mutually dependent, with an arbitrary joint distribution $P(E_1^n)$. We still require the $B_{ij}$ to be mutually independent and independent of all the $E_j$. Thus we are ready to define the model class:

**Definition 1** *A* noisy-or model with latent confounding *is a structural equation model over binary variables $X_1^n$, where the structural equations obey the form in Equation 2, the link variables $B_{ij}$, with link probabilities $b_{ij} = P(B_{ij} = 1) \in (0, 1)$, are mutually independent and independent of the disturbance terms $E_1^n$, but the disturbance terms $E_1^n$ can be mutually dependent with an arbitrary joint distribution $P(E_1^n)$.*

Such a model can represent any noisy-OR model with latent variables in which no latent confounders have observed parents. So given a set $X_I$ (with $I \subseteq$

$\{1, \ldots, n\}$) of variables that are subject to an intervention, the generating model assumes that first the values of $X_I$ are drawn from the chosen interventional distribution $\prod_{i \in I} P(X_i)$, the disturbances $E_1^n$ are drawn from $P(E_1^n)$, and the link variables $B_{ij}$ are drawn from $\prod_{ij} P(B_{ij})$. Subsequently, the values of the non-intervened observed variables $X_{\{j:\, j \notin I\}}$ are assigned using the noisy-OR formula (2) in accordance with the causal order of the variables.

## 3 IDENTIFIABILITY

In this section we show that both the structure and the full parametrization of any model in the space specified by Definition 1 is identified from the combination of a passive observational data set and a set of experiments where for each ordered pair of variables $(X_i, X_j)$ there is an experiment where $X_i$ is subject to an intervention and $X_j$ is not. (Note that such a set of experiments also satisfies the identifiability condition given by Hyttinen et al. (2010) for the linear case.) The proof proceeds in four steps: First we show that we can obtain a causal order over the variables from the set of experiments. Second, once a causal order has been established, we can determine the coefficients of edges connecting variables adjacent in the causal order. Third, we show how the remaining edges can be inferred by generalizing the formula of the previous step, and lastly we determine the distribution over the disturbance variables.

**Step 1.** We can determine a (partial) causal order of the set of variables by comparing for each ordered variable pair $(X_i, X_j)$ the probability $P(X_j = 1 \mid X_i = 1 \mid\mid X_I)$ with $P(X_j = 1 \mid X_i = 0 \mid\mid X_I)$, where $i \in I$ and $j \notin I$. In other words, we look at the conditional probability of $X_j = 1$ given the two possible values for $X_i$ in an experiment in which $X_i$ is subject to intervention and $X_j$ is passively observed. If these probabilities differ, $X_i$ must be an ancestor of $X_j$.[1] We then (arbitrarily) resolve the resulting partial order into a complete order over the variables and re-index the variables from the first in the causal order to the last by indices $1, \ldots, n$.

**Step 2.** For any two adjacent variables $X_i, X_{i+1}$ in the order, Glymour (1998) showed that the so-called "causal power" statistic due to Cheng (1997) can be used to determine the parameter $b_{i,i+1}$ of the (direct) causal influence of $X_i$ on $X_{i+1}$:

$$b_{i,i+1} = \mathrm{CP}(X_i \to X_{i+1} || X_I) \quad (6)$$

$$= \frac{P(X_{i+1} = 1 | X_i = 1 || X_I) - P(X_{i+1} = 1 | X_i = 0 || X_I)}{1 - P(X_{i+1} = 1 | X_i = 0 || X_I)}$$

where $i \in I$ and $i + 1 \notin I$. That is, the causal power of $X_i$ on $X_{i+1}$ can be computed as the difference in the conditional probabilities of $X_{i+1}$ given that $X_i$ is active or inactive in the experimental distribution intervening on $X_i$ (and possibly others, but not $X_{i+1}$). The denominator scales the difference in accordance with the baseline activation of $X_{i+1}$ in this distribution.

Glymour (1998), p. 53, conjectured that, similar to the linear case, trek-rules based on Cheng's causal power statistic could be used to identify the remaining causal structure of a noisy-OR model. Unfortunately this is not the case: The causal power of an intervened variable $X_i$ on a non-intervened variable $X_j$ is in general not equal to the sum-product of the edge parameters over all paths from $X_i$ to $X_j$. We thus proceed differently.

**Step 3.** The key to identifying all the remaining coefficients of edges "spanning" levels of the causal order lies in blocking all the *indirect* paths between the variables. This can be achieved by conditioning on all the intermediate variables in the causal order. We obtain a generalization of the formula in Step 2 for the coefficient of the direct causal influence of $X_i$ on $X_j$ with $i < j$, $i \in I$, and $j \notin I$:

$$b_{ij} = \mathrm{CP}(X_i \to X_j \mid X_{i+1}^{j-1} = 0 \mid\mid X_I) \quad (7)$$

The intervention on $X_i$ breaks any confounding of $X_i$ and $X_j$. The conditioning on $X_{i+1}^{j-1} = 0$ intercepts any *indirect* path $X_i \rightsquigarrow X_j$, *without* opening new dependencies via the latent variables, thanks to the context specific independence of the noisy-OR parameterization when conditioning on 0. As a result, $X_i$ is independent of the other causes of $X_j$, and so the CP-statistic provides a consistent estimate of $b_{ij}$.

Figure 3 illustrates the situation for a three variable model. On the left the direct causes of $X_3$ are $X_1$, $X_2$ and $E_3$. By intervening on $X_1$ we make $E_3$ independent of $X_1$ (Figure 3, right). By conditioning on $X_2 = 0$, $X_2$ no longer influences $X_3$, and $X_1$ does not influence $X_3$ indirectly through $X_2$. The context-specific independence guarantees that conditioning on $X_2 = 0$ does not destroy the independence of $X_1$ and $E_3$, because

$$(X_1 \perp\!\!\!\perp E_2 \mid\mid X_1) \;\Rightarrow\; (X_1 \perp\!\!\!\perp E_2 \mid X_2 = 0 \mid\mid X_1), \quad (8)$$

---

[1] Because we do not require faithfulness (Spirtes et al., 1993), effects through multiple paths might exactly cancel each other out. Additionally, some ancestral relationships may not always be detected when several variables are simultaneously intervened in an experiment. However, it is not difficult to see that each variable must be correlated with at least one direct effect. So, by transitivity of the ancestral relation, the set of experiments is sufficient to determine a valid partial order.

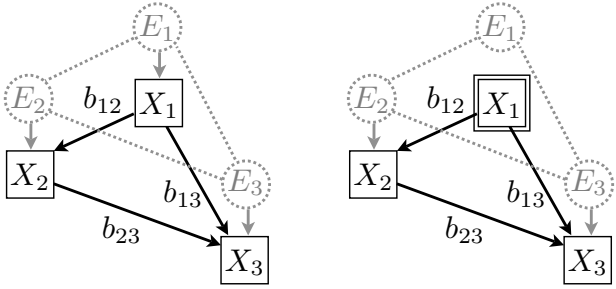

Figure 3: Left: full three variable model. Right: the same model after intervening on $X_1$, which breaks the influence of $E_1$. Any indirect paths $X_1 \rightsquigarrow X_3$ are intercepted when conditioning on $X_2 = 0$. The dotted edges represent the dependencies between the disturbances in the distribution $P(E_1^n)$.

which implies that $X_1 \perp\!\!\!\perp E_3 \mid X_2 = 0 \parallel X_1$. Consequently, all dependencies via the latent variables are blocked and $b_{13}$ is obtained as the corresponding causal power.

**Step 4.** That only leaves the probability distribution $P(E_1^n)$ over the disturbance variables: Given that the link probabilities among the observed variables are identified, we can use the passive observational distribution $P(X_1^n)$ to identify $P(E_1^n)$ as follows. The relationship between the two distributions is given by

$$P(X_1^n) = \sum_{E_1^n} P(X_1^n \mid E_1^n) P(E_1^n), \qquad (9)$$

which can be re-written in matrix notation as

$$\mathbf{p}_X = P_{X|E} \cdot \mathbf{p}_E, \qquad (10)$$

where the elements of the vector $\mathbf{p}_X$ are the probabilities of different configurations of the observed variables $X_1^n$, and the vector $\mathbf{p}_E$ contains the probabilities of the different disturbance configurations $E_1^n$.

The elements of the matrix $P_{X|E}$ can be written recursively as $P(X_1^n \mid E_1^n) = \prod_{j=1}^n P(X_j \mid X_1^{j-1}, E_j)$, in accordance with Equation 4. Consequently, each element of the matrix $P_{X|E}$ is a function of the already identified link probabilities. Hence we have a linear system of $2^n$ equations with $2^n$ unknowns. For instance, for three observed variables (as in Figure 3) we would have the matrix

$$P_{X|E} = \qquad (11)$$

$$\begin{array}{c|cccc}
X_1^3 \setminus E_1^3 & \cdots & (1,0,1) & (1,1,0) & (1,1,1) \\
\vdots & \ddots & \vdots & \vdots & \vdots \\
(1,0,1) & \cdots & (1-b_{12}) & 0 & 0 \\
(1,1,0) & \cdots & 0 & (1-b_{13})(1-b_{23}) & 0 \\
(1,1,1) & \cdots & b_{12} & b_{13}+b_{23}-b_{13}b_{23} & 1
\end{array}$$

where columns represent the different (joint) states of $E_1^3$ while rows represent the states of $X_1^3$ (for reasons of space we are only showing part of the matrix).

In order to show that $\mathbf{p}_E$ is identified, we have to show that $P_{X|E}$ is always invertible. Since the model definition implies that the disturbance variable $E_j$ deterministically makes $X_j = 1$ whenever $E_j = 1$, it is necessarily the case that $P(X_1^n \mid E_1^n) = 0$ whenever there is even a single $j$ such that $E_j = 1$ and $X_j = 0$. Hence, if we arrange the configurations of $X_1^n$ and $E_1^n$ in lexicographic order, as in (11), it is clear that the matrix $P_{X|E}$ is necessarily lower triangular. Furthermore, on the diagonal we have $E_1^n = X_1^n$ and so $P(X_1^n \mid E_1^n = X_1^n) = \prod_{j: X_j=0} \prod_{i \in \text{Pa}(j)} (1 - b_{ij})^{X_i}$. This is nonzero by the assumption in the model definition that all link probabilities $b_{ij}$ are strictly between 0 and 1. Lower-triangularity coupled with strictly nonzero values on the diagonal implies that the matrix is invertible, so we can always solve for the distribution $P(E_1^n)$.

Putting all of the above pieces together we have a constructive proof for the following theorem:

**Theorem 1** *All parameters of a noisy-OR model with latent confounding are identified from the combination of a passive observational data set and a set of experiments where for each ordered variable pair $(X_i, X_j)$ there is an experiment where $X_i$ is randomized and $X_j$ is observed.*

It is trivial to see that there are graphs for which the conditions on the set of experiments are necessary: If there are just two variables $X_1$ and $X_2$ and it is never the case that $X_1$ is subject to intervention while $X_2$ is passively observed, then the possible direct causal influence of $X_1$ on $X_2$ cannot be distinguished from a latent confounder of the two, and so the parameter $b_{12}$ of the edge $X_1 \to X_2$ cannot be consistently estimated without additional assumptions.

## 4 LEARNING

The four steps given in the previous section constitute a straightforward algorithm that – from a computational perspective – scales well to a large number of variables. However, if the sample size is limited then conditioning on all the intermediate variables $X_{i+1}^{j-1} = 0$ in (7) typically results in poor estimates, because there may be very few such samples available. In this section we give two learning algorithms that make better use of the available data, in order to learn the underlying structure and parameters of the model.[2]

---
[2]Code implementing both algorithms, and reproducing the simulations described in Section 6, is available at http://www.cs.helsinki.fi/u/ajhyttin/noisyor/

**Algorithm 1** EC-algorithm using efficient conditioning.

---

Determine the causal order of variables and re-index them as $X_1, \ldots, X_n$ according to Step 1 in Section 3. Initialize the graph over the variables to be empty.

For $X_i$ in $X_{n-1}, \cdots, X_1$

    For $X_j$ in $X_{i+1}, \cdots, X_n$

        Select the set $\mathcal{S}$ of experiments in which $X_i$ is subject to intervention.

        For each experiment in $\mathcal{S}$, find the smallest conditioning sets $X_C \subseteq \{X_{i+1}, \cdots, X_{j-1}\}$ such that all currently known directed paths $X_i \rightsquigarrow X_j$ are blocked (taking into account any edge deletions resulting from the experiment).

        For each experiment in $\mathcal{S}$, and each smallest conditioning set $X_C$, obtain preliminary estimates of the link probability using $\hat{b}_{ij} = CP(X_i \to X_j \mid X_C = 0 \parallel X_I)$.

        Combine the preliminary estimates to yield the final estimate of the link probability $b_{ij}$. If the link is statistically significant, add the edge $X_i \to X_j$ to the graph.

Calculate the least squares estimate for $P(E_1^n)$ from the equations $P(X_1^n \parallel X_I) = \sum_{E_1^n} P(X_1^n \mid E_1^n \parallel X_I) P(E_1^n)$.

---

**EC-algorithm** First, we observe that when the underlying causal structure is sparse, one can often avoid excessive conditioning. For instance, if we discover that there is no directed path from $X_1$ to $X_3$ through $X_2$, then we do not need to condition on $X_2 = 0$ when estimating $b_{13}$. Algorithm 1 generalizes this point and learns the links in such an order that the conditioning sets $X_C \subseteq X_{i+1}^{j-1}$ are the smallest ones necessary to block all indirect causal connections between the variables in question. Given that there may exist many such smallest sets, and many experiments from which such sets are obtained, we end up with many separate estimates of the same parameters, which can be combined to yield the final estimates of the link probabilities. In our implementation we use a simple weighted average for this combination, where the weights are given by the number of samples from which the preliminary estimates are computed. We have no principled way to test the aggregate of the (dependent) preliminary estimates underlying our final estimate, but we found that using a standard $\chi^2$-test ($\alpha = 0.01$) for the (conditional) independence of the pair of variables in question provides a good criterion for when to add an edge with the final link estimate to the model. If there are several candidate conditioning sets for the independence test, we trust the test for which the largest number of samples is available.

**EM-algorithm** The EC-algorithm still only uses part of the data when estimating each model parameter.

**Algorithm 2** Maximum likelihood estimation using the EM-algorithm.

---

Determine the causal order of variables and re-index them as $X_1, \ldots, X_n$ according to Step 1 in Section 3.

Let $N(X_1^n \parallel X_I)$ be the number of times $X_1^n$ is observed in an experiment that intervenes on $X_I$.

Pick initial values for the link probabilities $\{b_{ij}\}$ and the disturbance distribution $P(E_1^n)$, then iterate the following E- and M-steps until convergence:

E-step

    For each experiment calculate the probability of each disturbance configuration $E_1^n$ given each observed variable configuration $X_1^n$:

$$P^*(E_1^n | X_1^n \parallel X_I) = \frac{P(X_1^n | E_1^n, \{b_{ij}\} \parallel X_I) P(E_1^n)}{\sum_{E_1^n} P(X_1^n | E_1^n, \{b_{ij}\} \parallel X_I) P(E_1^n)}$$

M-step

    For each variable $X_j$ find the link probabilities $b_{\bullet j}$ for all incoming edges to node $X_j$ that maximize the expected log-likelihood

$$\sum_{\substack{X_I, X_1^n \\ E_1^n}} N(X_1^n \parallel X_I) P^*(E_1^n | X_1^n \parallel X_I) \log P(X_j | X_1^{j-1}, E_j, b_{\bullet j})$$

    Update the estimate of the distribution $P(E_1^n)$ by

$$P(E_1^n) = \frac{\sum_{X_I, X_1^n} N(X_1^n \parallel X_I) P^*(E_1^n | X_1^n \parallel X_I)}{\sum_{X_I, X_1^n, E_1^n} N(X_1^n \parallel X_I) P^*(E_1^n | X_1^n \parallel X_I)}$$

---

For instance, only experiments that intervene on $X_i$ contribute to the estimation of $b_{ij}$. Rather than ignoring the data from the other experiments, we can integrate it by maximizing the likelihood of the model. While the confounding makes direct optimization of the log-likelihood computationally difficult, we give in Algorithm 2 the details of an Expectation-Maximization algorithm (Dempster et al., 1977). In the E-step we estimate, for each experiment, the probability distribution of the unobserved disturbances $E_1^n$ given $X_1^n$, using the parameter settings of the previous round. In the M-step the maximization can then be performed with respect to the parameters associated with each node separately. Note that when a node has an arbitrary number of observed parents there is no known closed form solution for the ML-estimates of a noisy-OR conditional distribution, so we need to resort to iterative methods in the inner loop. Because of the exponential number of states of the unknown disturbances, the maximum likelihood approach is in practice restricted to relatively small models, but in such cases the EM-algorithm clearly outperforms the EC-algorithm in terms of accuracy (see Section 6). While the identifiability result of Section 3 guarantees that the MLE is consistent, we have not shown that the log-likelihood function is unimodal. Nevertheless, in our simulations the EM-algorithm always appeared to converge towards the true parameter values as we increased the sample size.

As is well known, maximizing the likelihood will in some cases (in particular when the amount of data is relatively limited) give parameter values that yield predictive probabilities exactly equal to zero. If this is undesirable, one can use suitable priors on the parameters and instead seek *maximum a posteriori* (MAP) parameter estimates, with only minimal changes to the EM-algorithm (Dempster et al., 1977).

## 5 NEGATIVE CAUSES

A significant restriction of the basic noisy-OR model is that all causes are necessarily *generative*, in the sense that if a parent is turned 'on' (i.e. is set to 1) – all other things being equal – this necessarily increases the probability that the child is 'on'. As an analogy, in a linear model this restriction would correspond to having only positive edge coefficients. While confounding makes it possible that a parent and child are nevertheless negatively correlated in the passive observational data, it would certainly be desirable to allow for variables having a *negative* causal effect on some of the other variables.

Fortunately, it is quite easy to generalize the noisy-OR model to allow for negative causes. We simply replace the basic model in (2) by

$$X_j := \bigvee_{i \in \mathrm{Pa}(j)} (B_{ij} \wedge \widetilde{X}_i) \vee E_j, \qquad (12)$$

where for generative causes $\widetilde{X}_i = X_i$ and for negative causes $\widetilde{X}_i = \neg X_i$. We distinguish the two types of causes by extending the parameter space of the link probabilities such that $b_{ij} \in (-1, 0) \cup (0, 1)$, with negative parameters representing negative causes. Such an extension of the basic noisy-OR model allows us to work with an essentially continuous parameter space, and it is not necessary to learn separate models with generative or negative causes (Neal, 1990).[3]

It is straightforward to see that we do not lose the context specific independence property when allowing such negative causes in the model. Hence we can still obtain identifiability, and Theorem 1 applies, for these generalized noisy-OR models in the following way. In steps 2 and 3 of Section 3, we only need to detect when a given link is negative, which is the case if and only if the corresponding causal power (CP) in (6) or (7) is negative. While the causal power in this case does not directly produce the (negative) $b_{ij}$, we need simply negate the parent $X_i$ and use the standard formula, obtaining $-b_{ij}$. Similarly, in step 4 of Section 3, we only need a trivial modification of the conditional probabilities in (4) to take account of the negative causes we found in the previous steps. Learning the extended model follows exactly the procedures of Section 4, and all of our simulations employ this full extended model.

---

[3]Note that if a link probability were to equal zero the corresponding parent simply does not affect the child, so such values are excluded in the model definition for identifiability reasons.

Finally, note that the negative causes in our formulation are not fully equivalent to 'preventive' causes as discussed in psychology and philosophy (Cheng, 1997). In those models a preventive cause can (when the corresponding link variable is on) *on its own* ensure that the effect variable is off. When a variable has more than one parent, the family of conditional probability distributions representable by such models is different from the family of those that can be represented with the generalization employed in this paper.

## 6 SIMULATIONS

Here we present three analyses of the performance of our algorithms and model formulations that address accuracy, scalability and robustness.

We start with a straightforward accuracy test of our algorithms. We generated data, of varying sample sizes, from 100 (generalized) noisy-OR models with latent confounding over eight variables (see Sections 2 and 5). To ensure that the choice of smallest conditioning sets in the EC-algorithm is relevant, we limited the number of parents to a maximum of two. The link probabilities were drawn from a uniform distribution on the interval $[0.1, 0.9]$, while the $2^n$ probability assignments of the (arbitrary) disturbance distribution were sampled using a Dirichlet distribution $\mathrm{Dir}(\mathbf{1})$. We used a sequence of eight experiments, each uniformly randomizing a different observed variable, to generate the data. This sequence, together with a passive observational data set, satisfies the identifiability conditions of Theorem 1. For each model we drew a *total* of $100 - 100{,}000$ samples: $1/9$ of the samples were collected in the passive observational setting and $1/9$ each were generated from the eight experiments. We used our three algorithms to learn the models: (i) ID, the algorithm based on the steps of the identifiability proof (Section 3); (ii) EC, the algorithm that efficiently selects conditioning sets (Algorithm 1); and (iii) EM, the algorithm that maximizes the likelihood (Algorithm 2).

The left plot of Figure 4 shows the accuracy of the estimated link probabilities among the observed variables.

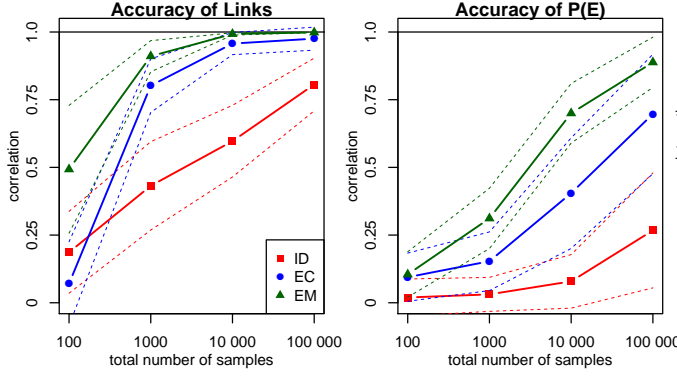

Figure 4: Accuracy of the learning algorithms. Each point on the solid lines is the average over 100 models. The dashed lines indicate ±1 standard deviation.

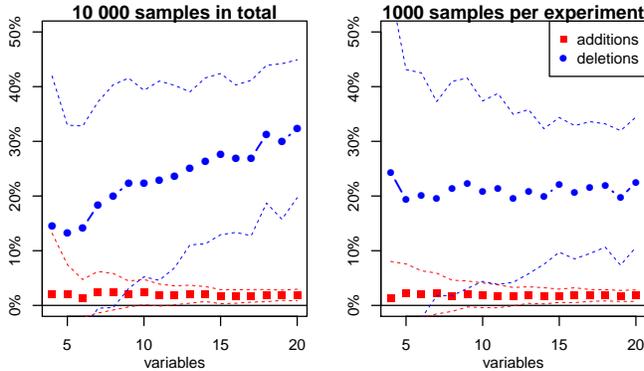

Figure 5: Structural errors when using the EC-algorithm. Each point on the solid lines is the average over 100 networks. The dashed lines indicate ±1 standard deviation.

We measured the linear correlation between the vector of estimated and true values of the link probabilities $b_{ij}$ for different total sample sizes for the three algorithms. In the right plot we used the same measure for the accuracy of the estimated vector of probability assignments of the disturbance distribution $P(E_1^n)$. As expected, selecting the smallest conditioning sets improves accuracy, while using the entire data in the EM-algorithm is better yet. The EM-algorithm achieves an accuracy of about 0.9 with only 1,000 total samples, i.e. with only 125 samples per experiment, for an 8-node graph. The estimates of the disturbance distribution are – unsurprisingly, given that $P(E_1^n)$ can take an arbitrary form – much less accurate.

For larger structures the EM-algorithm is computationally infeasible, but the EC-algorithm scales well. For each $n = 4 \ldots 20$ we generated 100 noisy-or models (without confounding) with $n + \lfloor n/2 \rfloor$ variables. To include confounding we considered the first $\lfloor n/2 \rfloor$ variables in the causal order as unobserved. For each model we performed $n$ experiments randomizing a sin-gle variable and used the EC-algorithm to learn the model. Figure 5 shows the accuracy of the estimates in terms of structural measures: the percentage of edge additions (false positive rate) and deletions (false negative rate). The number of edge reversals was negligible. On the left the models were learned on the basis of a *total* of 10,000 samples evenly divided over the $n$ experiments and a passive observational data set (no matter how large $n$ was), while for the right plot 1,000 samples were used per experiment (for a total of $1,000(n + 1)$ samples).

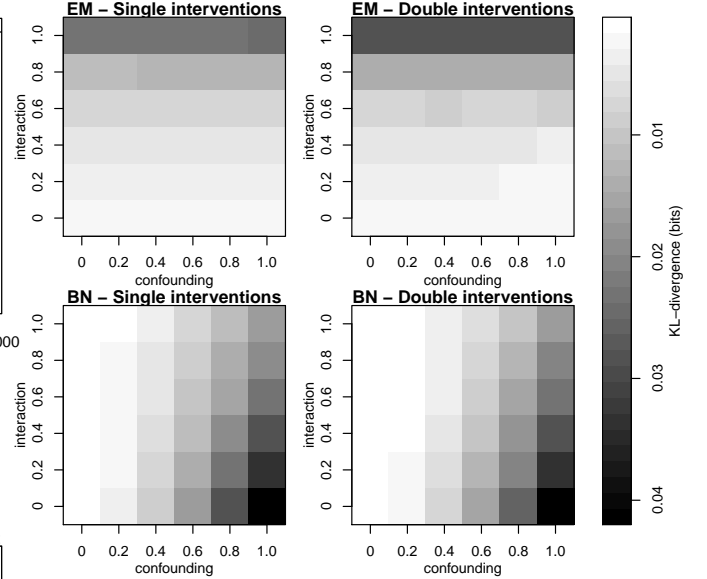

Figure 6: Predictive accuracy of the (generalized) noisy-OR model with latent confounding (EM) and Bayesian networks (BN). The shade of each square represents the average KL-divergence over all possible experiments randomizing one (left) or two (right) variables per experiment on 100 networks. Lighter shades indicate better results.

With a constant number of total samples (left plot), the percentage of deletions increases as we have fewer samples for the individual experiments. In contrast, with a constant number of samples per experiment, the rate remains approximately constant as the number of variables increases: about 20% of the edges present in the true model are deleted, the majority of which have low link probabilities or are edges for which after conditioning we only have a very small number of samples for estimation. As expected, the false positive rate (edge additions) is relatively constant at the p-value threshold (0.01) of the independence test (see Section 4).

Lastly, we explore the robustness of our algorithms when the data is generated by a model not satisfying our model space assumptions. One feature that can-not be represented by our model are interactive causal effects; and one way to generate such interaction is to add interaction terms into the structural equation of the generalized noisy-OR parameterization, analogous to the linear case:

$$X_k := \left[\bigvee_i (B_{ij} \wedge \widetilde{X}_i)\right] \vee \left[\bigvee_{i,j} (I^{++}_{ijk} \wedge X_i \wedge X_j)\right] \vee$$

$$\left[\bigvee_{i,j} (I^{+-}_{ijk} \wedge X_i \wedge \neg X_j)\right] \vee \left[\bigvee_{i,j} (I^{-+}_{ijk} \wedge \neg X_i \wedge X_j)\right] \vee$$

$$\left[\bigvee_{i,j} (I^{--}_{ijk} \wedge \neg X_i \wedge \neg X_j)\right] \vee E_k \quad (13)$$

where $i, j \in \text{Pa}(k)$, and $i \neq j$, and, e.g. $I^{+-}_{ijk}$ is the 'interaction variable' of the pairwise interaction term when $X_i = 1$ and $X_j = 0$, and $P(I^{+-}_{ijk} = 1) = p^{+-}_{ijk}$, with $p^{+-}_{ijk} \sim \text{Unif}[0, y]$ and $y \leq 1$. This is an extension of the (generalized) noisy-OR model in (12) to interactive effects.

The advantage of this "noisy i(nteractive)-OR" model is two-fold: By varying the maximum $y$ in the Uniform distribution regulating the probabilities of the interactive terms, we can control the amount of interaction in the model (note that when $y = 0$ then the model reverts to our earlier generalized noisy-OR model), and by varying the amount of dependence between the disturbances $E_i$, we can control the amount of confounding: We take an arbitrary distribution $P(E^n_1)$ over the disturbances to represent full confounding, and the product of its marginals, $\prod_i P(E_i)$, to represent no confounding. By varying $x$ from 0 to 1 in $xP(E^n_1) + (1-x)\prod_i P(E_i)$ we can construct any intermediate level of confounding.

For different levels of confounding (x-axis) and interaction (y-axis) we generated 100 5-variable models with the noisy i-OR-parameterization and sampled a total of 10,000 data points from 5 experiments randomizing a single variable and one passive observation. We compared the fit provided by our model (learned with the EM-algorithm) against that given by Bayesian learning of a Bayesian network (Heckerman et al., 1995) (MAP structure and parameters, for a standard BDeu prior with equivalent sample size set to 1; experiments were integrated using the procedure described by Cooper and Yoo (1999)). Note that a Bayes net without latents is essentially a completely general acyclic causal model over the observed variables, but does not allow for confounding.[4]

---
[4]Although the FCI-algorithm may at first pass seem like a more appropriate candidate for comparison (since it does not assume causal sufficiency), it only provides ancestral relationships. Here we are interested in distinguishing direct from indirect causes, and as we noted in Section 1 any procedure using only independence constraints would not be able to distinguish the structures in Figure 2.

In Figure 6 we show the average predictive accuracy, measured as KL-divergence, for experiments randomizing one variable (left column) or two variables (right column). Consistent with the different model space assumptions the results show that our noisy-OR model performs better when there is confounding but no interactive causation, the Bayesian networks perform better when there is interaction and no confounding, and there is not much difference when the model space assumptions fail for both.

# 7 CONCLUSION

We have specified a model for binary variables with latent confounding and have provided a condition on the set of experiments that is necessary and sufficient for model-identifiability when faithfulness is not assumed. This condition is similar to the corresponding linear case described in Hyttinen et al. (2010), but the learning algorithms here take a very different form. To our knowledge, this is the first approach that guarantees identifiability using experimental data with model space assumptions that are this weak. The simulations test the performance of our algorithms with regard to accuracy and scalability, and compare our procedures to standard Bayes net learning. Going forward, we intend to (i) investigate the use of stronger priors on the disturbance distribution to improve estimation performance (e.g. based on Markov random fields), (ii) provide a characterization of what can be learned when the identifiability conditions are not satisfied, (iii) extend the current approach to cyclic ('non-recursive') models and (iv) to consider various extensions to models with interactive causal effects.

### Acknowledgements

We thank three anonymous reviewers for helpful comments. A.H. & P.O.H. were supported by the Academy of Finland (project #1125272) and by University of Helsinki Research Funds (project #490012).

### References

Cheng, P. W. (1997). From covariation to causation: A causal power theory. *Psychological Review*, 104(2):367–405.

Claassen, T. and Heskes, T. (2010). Causal discovery discovery in multiple models from different experiments. In *NIPS 2010*.

Cooper, G. and Yoo, C. (1999). Causal discovery from a mixture of experimental and observational data. In *UAI 1999*.

Danks, D. and Glymour, C. (2001). Linearity properties of Bayes nets with binary variables. In *UAI 2001*.

Dempster, A., Laird, N., Rubin, D., et al. (1977). Maximum likelihood from incomplete data via the EM algorithm. *Journal of the Royal Statistical Society. Series B (Methodological)*, 39(1):1–38.

Eberhardt, F., Hoyer, P. O., and Scheines, R. (2010). Combining experiments to discover linear cyclic models with latent variables. In *AISTATS 2010*.

Glymour, C. (1998). Learning causes: Psychological explanations of causal explanation. *Minds and Machines*, 8:39–60.

Heckerman, D., Geiger, D., and Chickering, D. M. (1995). Learning Bayesian networks: The combination of knowledge and statistical data. *Machine Learning*, 20(3):197–243.

Hyttinen, A., Eberhardt, F., and Hoyer, P. O. (2010). Causal discovery for linear cyclic models. In *PGM 2010*.

Neal, R. M. (1990). Learning stochastic feedforward networks. Technical report, University of Toronto.

Pearl, J. (1986). Fusion, propagation, and structuring in belief networks. *Artificial Intelligence*, 29(3):241–288.

Pearl, J. (2000). *Causality: Models, Reasoning, and Inference*. Cambridge University Press.

Peng, Y. and Reggia, J. (1986). Plausibility of diagnostic hypotheses. In *AAAI 1986*.

Spirtes, P., Glymour, C., and Scheines, R. (1993). *Causation, Prediction, and Search*. Springer-Verlag.

Spirtes, P., Meek, C., and Richardson, T. (1995). Causal inference in the presence of latent variables and selection bias. In *UAI 1995*.

## A  DETAILS OF EXAMPLE 1

Here, we give the details of the example discussed in Section 1, illustrating that the combination of passive observational data and experiments intervening on a single variable at a time is in general not sufficient for full identification of the underlying model. Consider the model given in Figure 1. In the passive observational case we obtain

$$P(X_3 = 1 \mid X_1 = 0, X_2 = 0) = 0.5$$
$$P(X_3 = 1 \mid X_1 = 1, X_2 = 0) = 0.3$$
$$P(X_3 = 1 \mid X_1 = 0, X_2 = 1) = 0.3$$
$$P(X_3 = 1 \mid X_1 = 1, X_2 = 1) = 0.28,$$

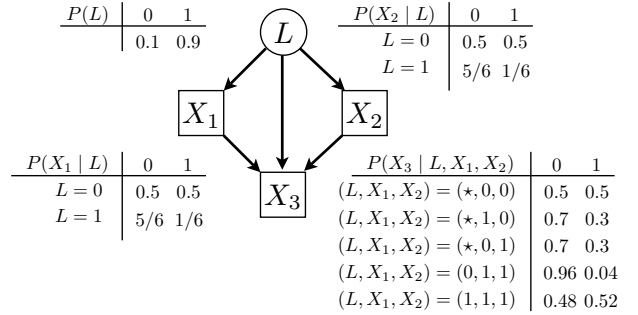

Figure 7: Alternative parameterization yielding the exact same distributions as Figure 1 for all experiments randomizing only one variable, and when passively observed.

and when intervening on $X_1$ (note the symmetry with respect to $X_1$ and $X_2$) we have

$$P(X_3 = 1 \mid X_1 = 0, X_2 = 0 \mid\mid X_1) = 0.5$$
$$P(X_3 = 1 \mid X_1 = 0, X_2 = 1 \mid\mid X_1) = 0.3$$
$$P(X_3 = 1 \mid X_1 = 1, X_2 = 0 \mid\mid X_1) = 0.3$$
$$P(X_3 = 1 \mid X_1 = 1, X_2 = 1 \mid\mid X_1) = 0.4,$$

while marginalizing out $X_2$ we obtain

$$P(X_3 = 1 \mid X_1 = 1 \mid\mid X_1) = 0.32$$
$$P(X_3 = 1 \mid X_1 = 0 \mid\mid X_1) = 0.46.$$

For the sake of argument, let us assume that we are to decide how to set $X_1$ and $X_2$, with the goal of maximizing the probability of obtaining $X_3 = 1$. Based on the above data, the natural choice seems to be to set $X_1 = 0$ and $X_2 = 0$. Yet, from the model specification, we know that setting $X_1 = 1$ and $X_2 = 1$ is the optimal choice:

$$P(X_3 = 1 \mid X_1 = 0, X_2 = 0 \mid\mid X_1, X_2) = 0.5$$
$$P(X_3 = 1 \mid X_1 = 1, X_2 = 1 \mid\mid X_1, X_2) = 0.6.$$

To formally prove that the model of Figure 1 is not identifiable from the combination of a passive observational dataset and experiments randomizing a single variable, we give in Figure 7 an alternative parametrization which yields the same passive observational distribution and single-intervention distributions, yet results in a different distribution for $X_3$ when simultaneously intervening on both $X_1$ and $X_2$. With the alternative model, inferences made on the basis of single-intervention experiments are correct: setting $X_1 = 0$ and $X_2 = 0$ is the optimal choice to maximize the probability of $X_3 = 1$, because for this model

$$P(X_3 = 1 \mid X_1 = 0, X_2 = 0 \mid\mid X_1, X_2) = 0.5$$
$$P(X_3 = 1 \mid X_1 = 1, X_2 = 1 \mid\mid X_1, X_2) = 0.472.$$